\documentclass[11pt,a4paper]{article}
\usepackage[hyperref]{acl2019}
\usepackage{times}
\usepackage{array}
\usepackage{tabularx}
\usepackage{latexsym}
\usepackage{graphicx}
\usepackage{amsmath}
\usepackage{url}
\usepackage{multirow}
\usepackage[inline]{enumitem}

\aclfinalcopy 
\def\aclpaperid{2739}

\newcommand{\PreserveBackslash}[1]{\let\temp=\\#1\let\\=\temp}
\newcolumntype{R}[1]{>{\PreserveBackslash\raggedleft}p{#1}}

\title{Improving Neural Conversational Models \\ with Entropy-Based Data Filtering}

\author{Richard Csaky \\
Department of Automation and Applied Informatics \\
Budapest University of Technology and Economics \\
\texttt{ricsinaruto@hotmail.com} \\\AND
Patrik Purgai \\
Department of Automation and Applied Informatics \\
Budapest University of Technology and Economics \\
\texttt{purgai.patrik@gmail.com} \\\And
Gabor Recski \\
Apollo.AI \\
\texttt{gabor@apollo.ai} \\}

\date{}

\begin{document}
\maketitle
\begin{abstract}
Current neural network-based conversational models lack diversity and generate
boring responses to open-ended utterances. {\it Priors} 
such as persona, emotion, or topic provide additional
information to dialog models to aid response generation, but annotating a
dataset with priors 
is expensive and such annotations are rarely available. While previous methods
for improving the quality of open-domain response generation focused on either
the underlying model or the training objective, we present a method of
filtering dialog datasets by removing generic utterances from training data
using a simple entropy-based approach that does not require human supervision.
We conduct extensive experiments with different variations of our method, and
compare dialog models across 17 evaluation metrics to show that training on
datasets filtered this way results in better conversational quality as
chatbots learn to output more diverse responses.
\end{abstract}

\section{Introduction}
\label{sec:introduction}
Current open-domain neural conversational models (NCM) are trained on
pairs of source and target utterances in an effort to maximize the likelihood
of each target given the source \cite{Vinyals:2015d}. However, real-world
conversations are much more complex, and a plethora of suitable targets
(responses) can be adequate for a given input. We propose a data filtering approach where the ``most open-ended'' inputs - determined by calculating the entropy of the distribution over target utterances - are excluded from the training set. 
We show that dialog models can be improved using this simple unsupervised method which can be applied to any conversational dataset. We conduct several experiments to uncover how some of the current open-domain dialog evaluation methods behave with respect to overfitting and random data. Our software for filtering dialog data and automatic evaluation using 17 metrics is released on GitHub under an MIT license\footnote{github.com/ricsinaruto/NeuralChatbots-DataFiltering}\footnote{github.com/ricsinaruto/dialog-eval}. This paper exists in poster\footnote{\url{https://bit.ly/2Myu060}} and blog post\footnote{\url{https://bit.ly/2T1JJMk}} form as well.

\section{Background}
\label{sec:background}
Most open-domain NCMs are based on neural network architectures developed
for machine translation (MT, \citet{Sutskever:2014,Cho:2014,Vaswani:2017}).
Conversational data differs from MT data in that targets to the same
source may vary not only grammatically but also semantically
\cite{Wei:2017,Tandon:2017}: consider plausible replies to the question
\textit{What did you do today?}. Dialog datasets also contain generic responses, 
e.g. \textit{yes}, \textit{no} and \textit{i don't know}, that appear in a large and diverse 
set of contexts \cite{Mou:2016,Wu:2018}. Following the approach of modeling 
conversation as a sequence 
to sequence (\texttt{seq2seq}, \citet{Sutskever:2014}) transduction of single
dialog turns, these issues can be referred to as the \textit{one-to-many},
and \textit{many-to-one} problem. \texttt{seq2seq} architectures are not suited 
to deal with the ambiguous nature of dialogs since they are 
inherently deterministic, meaning that once trained they cannot output 
different sequences to the same input. Consequently they tend to produce boring 
and generic responses \cite{Li:2016d,Wei:2017,Shao:2017b,Zhang:2018a,Wu:2018}.

Previous approaches to the \textit{one-to-many}, \textit{many-to-one} problem 
can be grouped into three categories. One approach involves feeding extra 
information to the dialog model such as dialog history 
\cite{Serban:2015,Xing:2018}, categorical information like persona 
\cite{Li:2016a,Joshi:2017,Zhang:2018}, mood/emotion \cite{Zhou:2018,Li:2017b}, 
and topic \cite{Xing:2017,Liu:2017,Baheti:2018}, or through knowledge-bases 
\cite{Dinan:2018,Ghazvininejad:2018,Zhu:2017,Moghe:2018}. A downside to these approaches 
is that they require annotated datasets which are not always available, or 
might be smaller in size. Augmenting the model itself, with e.g.~latent 
variable 
sampling \cite{Serban:2017b,Zhao:2017,Zhao:2018a,Gu:2019,Park:2018,Shen:2018,Gao:2019}, 
or improving the decoding process 
\cite{Shao:2017b,Kulikov:2018,Mo:2017,Wang:2018a} is also a popular approach. 
Sampling provides a way to generate more diverse responses, however such models 
are more likely to output ungrammatical or irrelevant 
responses. Finally, directly modifying the loss function \cite{Li:2016d}, or 
training by reinforcement 
\cite{Li:2016b,Serban:2017a,Li:2016c,Lipton:2017,Lewis:2017} or adversarial 
learning \cite{Li:2017a,Ludwig:2017,Olabiyi:2018,Zhang:2018b} has also been 
proposed, but this is still an open research problem, as it is far from trivial 
to construct objective functions that capture conversational goals better than 
cross-entropy loss.

Improving dataset quality through filtering is frequently used in the machine 
learning literature 
\cite{Sedoc:2018,Ghazvininejad:2018,Wojciechowski:2002} and data distillation methods in general are used both in machine translation and dialog systems \cite{Axelrod:2011,Li:2017c}. \citet{Xu:2018} introduced coherence for measuring the similarity between contexts and responses, and then filtered out pairs with low coherence. This improves datasets from a different aspect and could be combined with our present approach. However, natural conversations allow many adequate responses that are not similar to the context, thus it is not intuitively clear why filtering these should improve dialog models.
Our experiments 
also further support that cross-entropy is not an adequate loss function (shown qualitatively by \citet{Csaky:2017} and \citet{Tandon:2017}), 
by showing that many automatic metrics continue to improve after the 
validation loss reaches its minimum and starts increasing.
However, we found that the metrics steadily improve even after we can be certain that the model overfitted (not just according to the loss function). 
Further research is required, to determine whether this indicates that 
overfitted model responses are truly better or if 
it's a shortcoming of the metrics that they prefer such models.

Currently, there is no well-defined automatic evaluation method 
\cite{Liu:2016}, and while some metrics that correlate more with human 
judgment have been proposed recently \cite{Li:2017a,Lowe:2017,Tao:2018}, they 
are harder to measure than simpler automatic metrics like perplexity or BLEU 
\cite{Papineni:2002}. Furthermore, even human evaluation has its downsides, 
like high variance, high cost, and difficulty of replicating experimental setups
\cite{Zhang:2018,Tao:2018}. Some works resort to human evaluations 
\cite{Krause:2017a,Fang:2018}, others use automatic metrics only
\cite{Olabiyi:2018,Xing:2018a,Kandasamy:2017,Shalyminov:2018,Xu:2018}, and 
some use both \cite{Shen:2018a,Xu:2018a,Baheti:2018,Ram:2018}.
While extensive human evaluation of the methods presented here is left for future work, we do conduct an especially thorough automatic evaluation both at the validation loss minimum and of overfitted models. We believe our experiments also shed light on 
the limitations of frequently used automatic metrics.

\section{Methods}
\label{sec:methods}
\subsection{Intuition}
We approach the \textit{one-to-many}, \textit{many-to-one} problem from a relatively new perspective: instead of adding more 
complexity to NCMs, we reduce the complexity of the dataset by 
filtering out a fraction of utterance pairs that we assume are primarily 
responsible for generic/uninteresting responses. Of the 72 000 unique source 
utterances in the DailyDialog dataset (see Section~\ref{ssec:dataset} for 
details), 60 000 occur with a single target only. For these it seems 
straightforward to maximize the conditional probability $P(T|S)$, $S$ and $T$ 
denoting a specific source and target utterance. However, in the case of 
sources that appear with multiple targets (\textit{one-to-many}), models are 
forced to learn some ``average'' of observed responses \cite{Wu:2018}. 

The entropy of response distribution of an utterance \(s\) 
is a natural measure of the amount of ``confusion'' introduced by \(s\). For example, the context \textit{What did you do today?} has high entropy, since it is paired with many different responses in the data, but \textit{What color is the sky?} has low entropy since it's observed with few responses. The \textit{many-to-one} scenario can be similarly formulated, where a diverse set of source utterances are observed with the same target (e.g.~\textit{I don't know} has high entropy). While this may be a less prominent issue in training NCMs, we shall still experiment with excluding such generic targets, as dialog models tend to generate them frequently (see Section~\ref{sec:background}).

\subsection{Clustering Methods and Filtering}
\label{ssec:ic}
We refer with \textsc{identity} to the following entropy computation method.
For each source utterance $s$ in the dataset we calculate the entropy
of the conditional distribution $T|S=s$, i.e. given a dataset $D$ of source-target
pairs, we define the \textit{target entropy} of $s$ as
\begin{equation} \label{eq:target_entropy}
H_{\text{tgt}}(s, D) = - \sum_{(s, t_i)\in D} p(t_i|s)\log_2 p(t_i|s)
\end{equation}
Similarly, \textit{source entropy} of a target utterance is
\begin{equation} \label{eq:source_entropy}
H_{\text{src}}(t, D) = - \sum_{(s_i, t)\in D} p(s_i|t)\log_2 p(s_i|t)
\end{equation}
The probabilities are based on the observed relative frequency of utterance 
pairs in the data.

For the purposes of this entropy-based filtering, we considered the possibility 
of also including some form of similarity measure between utterances that would 
allow us to detect whether a set of responses is truly diverse, as in the case 
of a question like \textit{What did you do today?}, or diverse only on the surface, such as in the case of a question like \textit{How old are you?} (since answers to the latter are semantically close). Measuring the entropy of semantic clusters as opposed to individual utterances may improve our method by reducing data sparsity. For example \textit{How are you?} can appear in many forms, like \textit{How 
are you \textless name\textgreater?} (see 
Section~\ref{ssec:sent2vec_results}). While the individual forms have low entropy (because they have low frequency), we may decide to filter them all if together they form a high-entropy cluster.

To this end we performed the filtering based not only on the set of all 
utterances, as in the case of \textsc{identity}, but also on clusters of 
utterances established by clustering their vector representations using the Mean Shift algorithm \cite{Fukunaga:1975}. Source and target utterances are clustered separately. In the
\textsc{avg-embedding} setup the representation \(R(U)\) of utterance \(U\) is 
computed by taking the average word embedding weighted by the smooth inverse 
frequency \(R(U) = \frac{1}{|U|} \sum_{w\in U}\frac{E(w) \cdot 0.001}{0.001 + 
p(w)}\) of words \cite{Arora:2016a}, where \(E(w)\) and \(p(w)\) are the 
embedding and the probability\footnote{Based on the observed 
relative frequency 
in the data.} of word \(w\) respectively. We also experiment with 
\textsc{sent2vec}\footnote{\url{https://github.com/epfml/sent2vec}}, a more 
sophisticated sentence embedding approach, which can be thought of as an 
extension of word2vec to sentences \cite{Pagliardini:2018}.

The \textit{target entropy} of a source cluster \(c_s\) is
\begin{equation} \label{eq:target_cluster_entropy}
H_{\text{tgt}}(c_s, C) = - \sum_{c_i\in C} p(c_i|c_s)\log_2 p(c_i|c_s)
\end{equation}
where \(C\) is the set of all clusters and $p(c_i | c_s)$ is the conditional probability of observing an utterance from cluster $i$ after an utterance from cluster $s$. In the context of these methods, the entropy of an utterance will mean the entropy of its cluster. Note that \textsc{identity} is
a special case of this cluster-based entropy computation method, since in 
\textsc{identity} a ``cluster'' is comprised of multiple examples of 
one unique utterance. Thus a target cluster's entropy is computed similarly to 
Equation~\ref{eq:source_entropy}, but using clusters as in 
Equation~\ref{eq:target_cluster_entropy}.

Entropy values obtained with each of these methods were used to filter dialog data in three ways. The \textsc{source} approach filters utterance pairs in which the source utterance has high entropy,
\textsc{target} filters those with a high entropy target, and finally the \textsc{both} strategy filters all utterance pairs that are filtered by either \textsc{source} or \textsc{target}. Some additional techniques did not yield meaningful improvement and were excluded from further evaluation.
Clustering based on the Jaccard similarity of the bag of words of utterances only added noise to \textsc{identity} and resulted in much worse clusters than \textsc{sent2vec}.
Clustering single occurrences of each unique utterance (as opposed to datasets with multiplicity) lead to less useful clusters than when clustering the whole dataset, probably because it resulted in less weight being given to the frequent utterances that we want to filter out.
K-means proved inferior to the Mean Shift algorithm, which is a density-based clustering algorithm and seems to work better for clustering vectors of sentences.
Filtering stop words before clustering did not improve the quality of clusters, probably because many utterances that we want to filter out contain a large number of stop words.

\section{Data Analysis}
\label{sec:clustering}
\subsection{Dataset}
\label{ssec:dataset}
With 90 000 utterances in 13 000 dialogs, 
DailyDialog \cite{Li:2017b}, 
our primary dataset, is comparable in size with the Cornell Movie-Dialogs 
Corpus \cite{Danescu:2011}, but contains real-world conversations. Using the \textsc{identity} approach, about 87\% of 
utterances have 0 entropy (i.e. they do not appear with more than one target), 
5\% have an entropy of 1 (e.g. they appear twice, with different targets), remaining values rise sharply to 7.
This distribution is similar for source and target utterances.

\begin{figure}[!ht]
\centering
\includegraphics[width=0.47\textwidth]{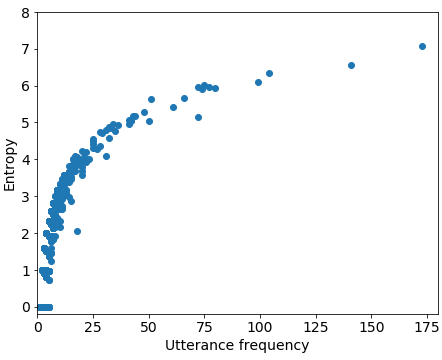}
\caption{Entropy of source utterances (computed with \textsc{identity}) with respect to utterance frequency.}
\label{fig:frequencies}
\end{figure}

\begin{figure}[!ht]
\centering
\includegraphics[width=0.47\textwidth]{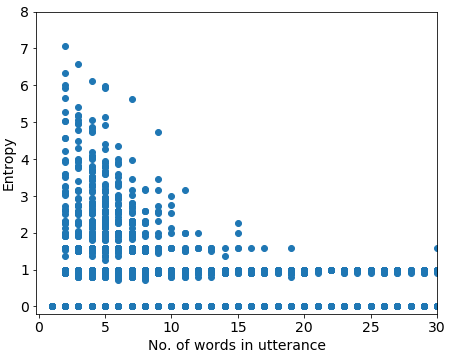}
\caption{Entropy of source utterances (computed with \textsc{identity}) with respect to utterance length.}
\label{fig:length}
\end{figure}

Entropy is clearly proportional to utterance frequency (Figure~\ref{fig:frequencies}),
but has a wide range of values among utterances of equal frequency.
For example, utterances with a frequency of 3 can have entropies ranging from 0 to
$\log_2 3\approx1.58$, the latter of which 
would be over our filtering threshold of 1 (see Section~\ref{ssec:parameters} 
for details on selecting thresholds). Since high-entropy 
utterances are relatively short, we also examined the relationship between entropy and utterance length (Figure~\ref{fig:length}). Given the relationship between frequency and entropy, it comes as no surprise that longer utterances have lower entropy.

\subsection{Clustering Results}
\label{ssec:sent2vec_results}
Compared to \textsc{identity}, both \textsc{sent2vec} and 
\textsc{avg-embedding} produce a much lower number of clusters with 0 entropy, 
but also a huge cluster with more than 5000 elements (the size of the second largest cluster
is below 500), which we didn't filter since it clearly doesn't group utterances 
with similar meaning. Generally, clusters were formed of similar utterances with 
the occasional exception of longer outlier utterances clustered together  
(instead of creating a separate cluster for each outlier), which can be 
attributed to the nature of the clustering algorithm. Overall, 
\textsc{sent2vec} appeared to produce better clusters than 
\textsc{avg-embedding}, as reflected in the evaluation in 
Section~\ref{sec:filt_experiments}.

\begin{figure}[!ht]
\centering
\includegraphics[width=0.47\textwidth]{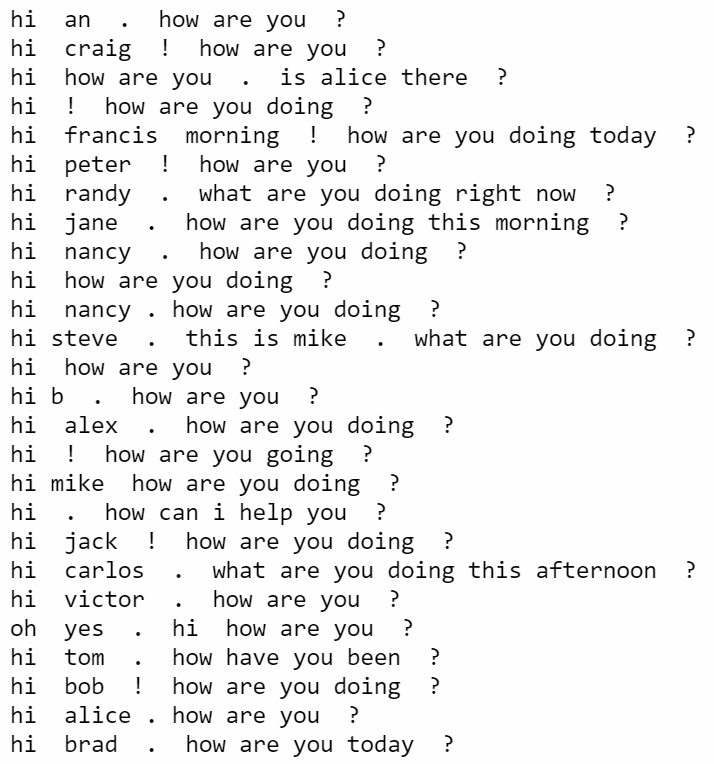}
\caption{A cluster produced by \textsc{sent2vec}.}
\label{fig:cluster_examples}
\end{figure}

We experimented with different bandwidth values\footnote{Bandwidth is like a 
radius in the latent space of utterance representations \cite{Fukunaga:1975}.} for the Mean Shift algorithm to produce clusters with as many elements as possible while also keeping the elements semantically similar. In an example cluster (Figure~\ref{fig:cluster_examples}) we can see that the clustering was able to group together several variants of \textit{How are you?}, 
in particular, those with different names. In general, we noticed that both in the case of \textsc{identity} and the clustering methods, utterances labeled with the highest entropy are indeed those generic sources and replies which we hoped to eliminate. See Appendix~\ref{ssec:high_entropy} for a selection of high entropy utterances and clusters.

\section{Experiments}
\label{sec:filt_experiments}
In this section the model and parameter setups are presented along with 17
evaluation metrics. Limitations of these metrics are discussed and 
a comparison between our filtering methods is presented on DailyDialog (Section~\ref{ssec:results}), and other datasets (Section~\ref{ssec:other_datasets}).

\subsection{Model and Parameters}
\label{ssec:parameters}
\begin{table}[h!]
\small
\begin{center}
 \begin{tabular}{p{1.4cm}p{0.5cm}p{0.5cm}p{1cm}p{0.9cm}p{0.8cm}}
 	Dataset &Type & Th. & \textsc{source} & \textsc{target} & \textsc{both} \\ \hline
 	\multirow{3}{*}{\bf DailyDialog}& \textsc{id} & 1 & 5.64\% & 6.98\% & 12.2\% \\
 	&\textsc{ae} & 3.5 & 5.39\% & 7.06\% & 12.0\% \\
 	&\textsc{sc} & 3.5 & 6.53\% & 8.45\% & 14.3\% \\
 	
 	\bf Cornell & \textsc{id} & 4 & - & 7.39\% & 14.1\% \\
 	\bf Twitter & \textsc{id} & 0.5 & - & 1.82\% & 9.96\% \\
 	
 \end{tabular}
\end{center}
\caption{\label{table:thresholds} Entropy threshold (Th.) and amount of data filtered for all datasets in the 3 filtering scenarios. \textsc{id} stands for \textsc{identity}, \textsc{ae} stands for \textsc{avg-embedding}, and \textsc{sc} for \textsc{sent2vec}.}
\end{table}

We use \texttt{transformer} \cite{Vaswani:2017} as our dialog model, an
encoder-decoder architecture relying solely on attention mechanisms 
\cite{Bahdanau:2015}. \texttt{transformer} 
has already been applied to a plethora of natural language processing tasks, 
including dialog modeling \cite{Dinan:2018,Mazare:2018,Devlin:2018}. We used the
official 
implementation\footnote{\url{https://github.com/tensorflow/tensor2tensor}} (see 
Appendix~\ref{ssec:parameters_appendix} for a report of hyperparameters). The vocabulary for DailyDialog was limited to 
the most frequent 16~384 words, and train / validation / test splits 
contained 71~517 / 9~027 / 9~318 examples, respectively.
\paragraph{Clustering and Filtering.} For \textsc{avg-embedding}
fastText\footnote{\url{https://fasttext.cc/}} embeddings were used. The 
bandwidth of Mean Shift was set to 0.7 and 3.5 for \textsc{avg-embedding} and 
\textsc{sent2vec}, which produced 40~135 and 23~616 clusters, respectively. Entropy thresholds and amount of data filtered can be found in Table~\ref{table:thresholds}. Generally we set the threshold so that filtered data amount is similar to the DailyDialog \textsc{identity} scenario. We also
set a threshold for the maximum average utterance length (15 and 20 for 
\textsc{avg-embedding} and \textsc{sent2vec}) in clusters that we considered 
for filtering, excluding outliers from the filtering process (see 
Section~\ref{ssec:sent2vec_results}).
\paragraph{Training and Decoding.} Word embeddings of size 512 were randomly 
initialized, batch size was set to 2048 
tokens, and we used the Adam optimizer \cite{Kingma:2014}. We experimented with 
various beam sizes \cite{Graves:2012b}, but greedy decoding performed better 
according to all metrics, also observed previously
\cite{Asghar:2017,Shao:2017b,Tandon:2017}.

\subsection{Evaluation Metrics}
\label{ssec:metrics}
As mentioned in Section~\ref{sec:background}, automatic evaluation of chatbots 
is an open research problem. In order to get as complete a picture as possible, 
we use 17 metrics that have been applied to dialog models over the past years, briefly described below. These metrics assess different aspects of response quality, thus models should be compared on the whole set of metrics.

\paragraph{Response length.} Widely used as a simple engagement indicator \cite{Serban:2017b,Tandon:2017,Baheti:2018}.
\paragraph{Word and utterance entropy.} The per-word entropy 
\(H_w = -\frac{1}{|U|} \sum_{w\in U}\log_2{p(w)}\) 
of responses is measured to determine their non-genericness \cite{Serban:2017b}.
Probabilities are calculated based on frequencies observed in the training data. We introduce the bigram version of this metric, to measure diversity at the bigram level as well. Utterance entropy is the product of \(H_w\) and \(|U|\), also reported at the bigram level.
\paragraph{KL divergence.} We use the KL divergence between model and 
ground truth (GT) response sets to measure how well a model can approximate the 
GT distribution of words. Specifically, we define distributions \(p_{gt}\) and \(p_m\)
based on each set of responses and calculate the KL divergence 
\(D_{kl} = \frac{1}{|U_{gt}|} \sum_{w\in U_{gt}} \log_2{\frac{p_{gt}(w)}{p_{m}(w)}}\)
for each GT response. The bigram version of this metric is also reported.
\paragraph{Embedding metrics.} Embedding \textit{average}, \textit{extrema}, 
and \textit{greedy} are widely used metrics \cite{Liu:2016,Serban:2017b,Zhang:2018b}. 
\textit{average} measures the cosine similarity between the averages of word vectors of response and target utterances. \textit{extrema} constructs a 
representation by taking the greatest absolute value for each dimension among the word vectors in the response and target utterances and measures the cosine similarity between them. Finally, \textit{greedy} matches each response token to a target token (and vice versa) based on the cosine similarity between their embeddings and averages the total score across all words. For word embeddings and average word embedding representations, we used the same setup as in \textsc{avg-embedding}.
\paragraph{Coherence.} We measure the cosine similarity between pairs of input and response \cite{Xu:2018}. Although a coherence value of 1 would indicate that input and response are the same,
generally a higher value seems better as model responses tend to have lower coherence than targets.
\paragraph{Distinct metrics.} \textit{Distinct-1} and \textit{distinct-2} are widely used in the literature \cite{Li:2016d,Shen:2018a,Xu:2018}, measuring the ratio of unique 
unigrams/bigrams to the total number of unigrams/bigrams in a set of responses. 
However, they are very sensitive to the test data size, since increasing the number of examples in itself lowers their value. While the number of total words increases linearly, the number of unique words is limited by the vocabulary, and we found that the ratio decreases even in human data (see Appendix~\ref{ssec:eval_appendix} for details). It is therefore important to only compare \textit{distinct} metrics computed on the same test data.
\paragraph{Bleu.} Measuring n-gram overlap between response and target is widely used in the machine learning and dialog literature \cite{Shen:2018a,Xu:2018}. We report BLEU-1, BLUE-2, BLEU-3, and BLEU-4 computed with the 4th smoothing algorithm described in \citet{Chen:2014d}.

\begin{figure}[!ht]
\centering
\includegraphics[width=0.47\textwidth]{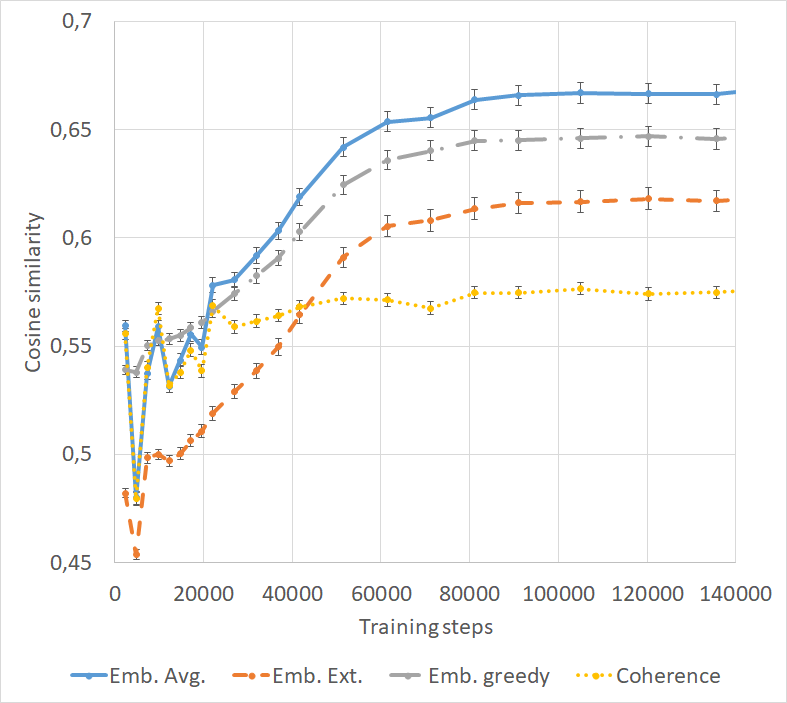}
\caption{Embedding metrics and coherence (on validation data) as a function of the training evolution of \texttt{transformer} on unfiltered data (DailyDialog).}
\label{fig:embedding_metric}
\end{figure}
\begin{figure}[!ht]
\centering
\includegraphics[width=0.47\textwidth]{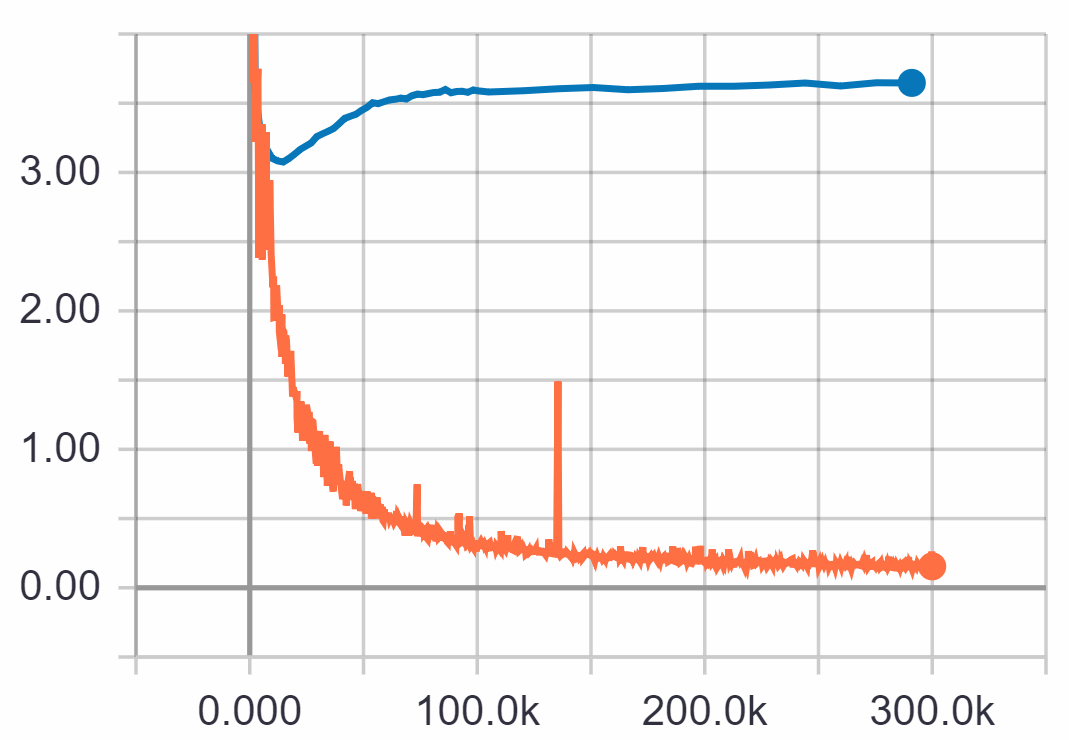}
\caption{Training (bottom) and validation (top) loss with respect to training steps of \texttt{transformer} trained on unfiltered data (DailyDialog).}
\label{fig:loss_base}
\end{figure}

\begin{table*}[h!]
\begin{center}
 \small
 \renewcommand{\arraystretch}{1.2}
 \begin{tabular}{p{0.1cm}p{0.1cm}R{0.6cm}p{0.4cm}p{0.4cm}p{0.4cm}R{0.5cm}p{0.4cm}R{0.6cm}p{0.4cm}p{0.4cm}p{0.4cm}p{0.4cm}p{0.5cm}p{0.4cm}p{0.4cm}p{0.4cm}p{0.4cm}p{0.4cm}}
 	
 	&& \(|U|\) & \(H_{w}^{u}\) & \(H_{w}^{b}\) & \(H_{u}^{u}\) & \(H_{u}^{b}\) & \(D_{kl}^{u}\) & \(D_{kl}^{b}\) & \textsc{avg} & \textsc{ext} & \textsc{gre} & \textsc{coh} & d1 &d2&b1&b2&b3&b4\\ \hline
 	
 	\multicolumn{2}{c}{\bf\textsc{trf}}&8.6 &7.30 &12.2 &63.6 &93 &.330 &.85&.540 &.497 &.552 &.538 &\bf.0290&.149&.142&.135&.130&.119 \\ \hline
 	
 	\multirow{3}{*}{\bf \rotatebox{90}{ID}} &\textsc{b}&9.8 &7.44 &12.3 &71.9 &105 &.315 &.77 &.559 &\bf.506&\it.555&.572 &.0247&.138&.157&.151&.147&.136 \\
 	&\textsc{t}&\it10.9&\bf7.67&\bf12.7&\bf83.2&\bf121&\bf.286&\bf.72&\bf.570&\bf.507&.554 &\bf.584&.0266&\bf.150&\bf.161&\bf.159&\bf.156&\bf.146\\
 	&\textsc{s}&9.4&7.19&11.9&66.4&98&.462&1.08&.540 &.495 &.553 &.538 &.0262&.130&.139&.133&.128&.117\\ \hline
 	
 	\multirow{3}{*}{\bf \rotatebox{90}{AE}} &\textsc{b}&7.9&7.25&12.0 &57.7 &83 &.447 &1.05&.524 &.486 &.548 &.524 &.0283&.132&.128&.121&.115&.105 \\
 	&\textsc{t}&8.6 &7.26 &12.1 &61.4 &90 &.425 &1.12&.526 &.492 &.548 &.529 &.0236&.115&.133&.127&.121&.111 \\
 	&\textsc{s}&\it9.0&7.21&11.9&\it65.1&\it95&.496&1.16&.536 &.490 &.548 &.538 &.0232&.109&.134&.130&.126&.116\\ \hline
 	
 	\multirow{3}{*}{\bf \rotatebox{90}{SC}} &\textsc{b}&10.0 &7.40 &12.3 &72.6 &108 &.383 &.97&.544 &.497 &.549 &.550 &.0257&.131&.145&.142&.138&.128 \\
 	&\textsc{t}&\bf11.2&\it7.49&\it12.4&\bf82.2&\bf122&.391 &.97&\it.565&\it.500&.552 &\it.572&.0250&.132&\it.153&\it.153&\it.152&\it.142 \\
 	&\textsc{s}&\bf11.1&7.15 &11.9 &74.4 &114 &.534 &1.27&.546 &\it.501&\bf.560&.544 &.0213&.102&.144&.139&.135&.125 \\
 \end{tabular}
\end{center}
\caption{\label{table:entropy_metrics_normal} Metrics computed at the minimum of the validation loss on the unfiltered test set (DailyDialog). \textsc{trf} refers to \texttt{transformer}, \textbf{ID} to \textsc{identity}, \textbf{AE} to \textsc{avg-embedding}, and \textbf{SC} to \textsc{sent2vec}. \textsc{source}-side, \textsc{target}-side, and filtering \textsc{both} sides are denoted by initials. Best results are highlighted with bold and best results separately for each entropy computing method are in italic (and those within a 95\% confidence interval).}
\end{table*}

\begin{table*}[h!]
\begin{center}
 \small
 \renewcommand{\arraystretch}{1.2}
 \begin{tabular}{p{0.1cm}p{0.1cm}p{0.5cm}p{0.4cm}p{0.4cm}R{0.5cm}p{0.4cm}p{0.5cm}p{0.5cm}p{0.5cm}p{0.4cm}p{0.4cm}p{0.4cm}p{0.5cm}p{0.4cm}p{0.4cm}p{0.4cm}p{0.4cm}p{0.4cm}}
 	
 	&& \(|U|\) & \(H_{w}^{u}\) & \(H_{w}^{b}\) & \(H_{u}^{u}\) & \(H_{u}^{b}\) & \(D_{kl}^{u}\) & \(D_{kl}^{b}\) & \textsc{avg} & \textsc{ext} & \textsc{gre} & \textsc{coh} & d1 &d2&b1&b2&b3&b4\\ \hline
 	
 	\multicolumn{2}{c}{\bf\textsc{trf}}&11.5 &7.98 &\bf13.4 &95 &142 &.0360 &.182&.655 &\bf.607 &\bf.640 &.567 &\bf.0465&\bf.297&\bf.333&.333&.328&.315 \\ \hline
 	
 	\multirow{3}{*}{\bf \rotatebox{90}{ID}} &\textsc{b}&\bf13.1&\bf8.08&\bf13.6 &\bf107&\bf162&.0473 &.210&\bf.668&\bf.608 &\bf.638 &\bf.598&.0410&.275&\bf.334&\bf.340&\bf.339&\bf.328 \\
 	&\textsc{t}&12.2 &8.04 &\bf13.6 &100 &150 &\bf.0335&\bf.181&\bf.665&\bf.610 &\bf.640 &.589 &.0438&.289&\bf.338&\bf.341&\bf.339&\bf.328\\
 	&\textsc{s}&12.3 &7.99 &\bf13.5 &101 &153 &.0406 &.187&.662 &\bf.610 &\bf.641 &.578 &.0444&.286&\bf.339&\bf.342&\bf.338&\bf.326 \\ \hline
 	
 	\multirow{3}{*}{\bf \rotatebox{90}{AE}} &\textsc{b}&11.9 &7.98 &\bf13.5 &98 &147 &.0395 &.197&.649 &.600 &.628 &.574 &.0434&.286&.318&.321&.318&.306 \\
 	&\textsc{t}&\it12.5&7.99 &\bf13.5 &\it102&\it155&.0436 &.204&.656 &.602 &.634 &\it.580&.0423&.279&.324&.327&.325&.313 \\
 	&\textsc{s}&12.1 &7.93 &\bf13.4 &99 &148 &.0368 &.186&.658 &.605 &\bf.636 &\it.578&.0425&.278&.325&.328&.324&.311 \\ \hline
 	
 	\multirow{3}{*}{\bf \rotatebox{90}{SC}} &\textsc{b}&12.8 &\bf8.07&\bf13.6 &105 &159 &.0461 &.209&.655 &.600 &.629 &\it.583&.0435&.282&.322&.328&.327&.316 \\
 	&\textsc{t}&\bf13.0&\bf8.06&\bf13.6 &\bf107&\bf162&.0477 &.215&.657 &.602 &.632 &\it.585&.0425&.279&.324&.330&.329&.318 \\
 	&\textsc{s}&12.1 &7.96 &\bf13.4 &100 &150 &\it.0353&.183&.657 &\bf.606 &\bf.638 &.576 &.0443&.286&.331&.333&.329&.317 \\ \hline
 	\multicolumn{2}{c}{\bf\textsc{RT}}& 13.5 &8.40 &14.2 &116 &177 &.0300 &.151&.531 &.452 &.481 &.530 &.0577&.379&.090&.121&.130&.125\\
 	\multicolumn{2}{c}{\bf\textsc{GT}}&14.1 &8.39 &13.9 &122 &165 &0 &0&1 &1 &1 &.602 &.0488&.362&1&1&1&1\\
 	
 \end{tabular}
\end{center}
\caption{\label{table:entropy_metrics_overfit} Metrics computed on the unfiltered test set (DailyDialog) after 150 epochs of training. \textsc{trf} refers to \texttt{transformer}, \textbf{ID} to \textsc{identity}, \textbf{AE} to \textsc{avg-embedding}, and \textbf{SC} to \textsc{sent2vec}. \textsc{source}-side, \textsc{target}-side, and filtering \textsc{both} sides are denoted by initials. Best results are highlighted with bold and best results separately for each entropy computing method are in italic (and those within a 95\% confidence interval). \textbf{GT} refers to ground truth responses and \textbf{RT} refers to randomly selected responses from the training set.}
\end{table*}

Normally metrics are computed at the validation loss minimum of a model, however
in the case of chatbot models loss may not be a good indicator of response quality (Section~\ref{sec:background}), thus we also looked at how our metrics progress during training. 
Figure~\ref{fig:embedding_metric} shows how coherence and the 
3 embedding metrics saturate after about 80-100k steps, and never decrease (we
ran the training for 300k steps, roughly 640 epochs).
Most metrics show a similar trend of increasing until 100k steps, and 
then stagnating (see Appendix~\ref{ssec:eval_appendix} for more figures).

In contrast, validation loss for the same training reaches its minimum after about 10-20k steps (Figure~\ref{fig:loss_base}). This again suggests the inadequacy of the loss function, but it also questions the validity of these metrics, as they seem to favor a model that overfitted the training data, which we can assume after 640 epochs. This could be due to the many identical inputs in train and test splits, 
because of the nature of dialog data. Most interesting are embedding metrics and BLEU scores (Section~\ref{ssec:results}), since they show that even after overfitting responses do not get farther from targets. This is in line with other findings reporting that qualitatively responses are better after overfitting 
\cite{Csaky:2017,Tandon:2017}, however occasionally they also tend to be too specific and irrelevant. We leave it for future work to conduct human evaluation between non-overfitted and overfitted models to solidify these claims. In light of these issues, we compare trainings on the DailyDialog dataset both at the validation loss minimum and at an overfitted point (150 epochs).

\subsection{DailyDialog Results}
\label{ssec:results}

\begin{table*}[h!]
\small
\begin{center}
 \begin{tabular}{p{6.5cm}p{8.5cm}}
 	Input & Response \\ \hline
 	\multirow{4}{6cm}{your starting salary is 2500 yuan a month and after you become a permanent employee it will be higher .} & \textbf{\textsc{base}:} i can tell you what is the best way to find a job . \\ 
 	& \textbf{\textsc{base-o}:} do you know what it is ? \\ 
 	& \textbf{\textsc{target}:} i 'm very interested in the position . \\ 
 	& \textbf{\textsc{target-o}:} that 's very nice of you . i 'll have to think about it . \\ \hline
 	
 	\multirow{4}{6cm}{you can greatly improve your grades by always reflecting on what you can improve on .} & \textbf{\textsc{base}:} i want to be a great singer . \\ 
 	& \textbf{\textsc{base-o}:} i really appreciate it . thanks . \\ 
 	& \textbf{\textsc{target}:} i think i am experienced in that position . i think i would like to make a good impression . \\ 
 	& \textbf{\textsc{target-o}:} i always liked it . thank you . \\ \hline
 	
 	\multirow{4}{6cm}{umm i think i would be hard to find something in that range in this area . you know the environment in the south district is the nicest .} & \textbf{\textsc{base}:} what about the kitchen ? \\ 
 	& \textbf{\textsc{base-o}:} what about the kitchen ? \\ 
 	& \textbf{\textsc{target}:} what about the kitchen ? \\ 
 	& \textbf{\textsc{target-o}:} what about the kitchen ? \\ \hline
 	
 \end{tabular}
\end{center}
\caption{\label{table:examples} Example inputs and responses from DailyDialog. \textsc{base} is trained on unfiltered data, and \textsc{target} is the model trained on \textsc{identity}, \textsc{target} filtered data. Models marked with \textsc{o} are evaluated at an overfitted point.}
\end{table*}

We compute metrics on the unfiltered test set to show that filtered 
trainings perform better even on utterances that would have been filtered from 
the training data. \textsc{trf}, the baseline
\texttt{transformer} model trained on unfiltered data is compared to the 9 trainings on filtered data. In all tables the 17 metrics from left to right are: response length, 
unigram and bigram entropy, unigram and bigram utterance entropy, unigram and 
bigram KL divergence, embedding \textit{average}, \textit{extrema} and 
\textit{greedy}, coherence, \textit{distinct-1} and \textit{distinct-2}, and finally, BLEU-1, BLEU-2, BLEU-3 and BLEU-4 (see Section~\ref{ssec:metrics}).

Evaluating at the minimum validation loss 
(Table~\ref{table:entropy_metrics_normal}) clearly shows that models 
trained on data filtered by \textsc{identity} and \textsc{sent2vec} are better 
than the baseline. \textsc{identity} 
performs best among the three methods, surpassing the baseline on all but the
\textit{distinct-1} metric. 
\textsc{sent2vec} is a close second, getting higher values on fewer metrics 
than \textsc{identity}, but mostly improving on the baseline. Finally, 
\textsc{avg-embedding} is inferior to the baseline, as it didn't produce clusters as meaningful as 
\textsc{sent2vec}, and thus produced a lower quality training set. It 
seems like filtering high 
entropy targets (both in the case of \textsc{identity} and \textsc{sent2vec}) 
is more beneficial than filtering sources, and 
\textsc{both} falls mostly in the middle as expected, since it combines the two filtering types. By removing example responses that are boring and generic from the dataset the model learns to improve response quality. Finding such utterances is useful for a number of purposes, but earlier it has been done mainly manually \cite{Li:2016b,Shen:2017}, whereas we provide an automatic, unsupervised method of detecting them based on entropy.

Every value is higher after 150 epochs of training than at the validation loss 
minimum (Table~\ref{table:entropy_metrics_overfit}). The most striking change is in the unigram KL divergence, which is now an order of magnitude lower.
\textsc{identity} still performs best, falling behind the 
baseline on only the two \textit{distinct} metrics. Interestingly this time \textsc{both} 
filtering was better than the \textsc{target} filtering. \textsc{sent2vec} 
still mostly improves the baseline and \textsc{avg-embedding} now
also performs better or at least as good as the baseline on most metrics. 
In some cases the best 
performing model gets quite close to the ground truth performance. On metrics that evaluate utterances without context (i.e. entropy, 
divergence, \textit{distinct}), randomly selected responses achieve similar values as the ground truth, 
which is expected. However, on embedding metrics, coherence, and BLEU,
random responses are significantly worse than those of any model evaluated.

\begin{table*}[h!]
\begin{center}
 \small
 \renewcommand{\arraystretch}{1.1}
 \begin{tabular}{p{0.1cm}p{0.1cm}R{0.6cm}p{0.4cm}p{0.4cm}R{0.5cm}R{0.5cm}R{0.6cm}R{0.6cm}p{0.3cm}p{0.3cm}p{0.3cm}p{0.4cm}p{0.8cm}p{0.8cm}p{0.3cm}p{0.3cm}p{0.3cm}p{0.3cm}}
 	
 	&& \(|U|\) & \(H_{w}^{u}\) & \(H_{w}^{b}\) & \(H_{u}^{u}\) & \(H_{u}^{b}\) & \(D_{kl}^{u}\) & \(D_{kl}^{b}\) & \textsc{avg} & \textsc{ext} & \textsc{gre} & \textsc{coh} & d1 &d2&b1&b2&b3&b4\\ \hline
 	
 	\multicolumn{2}{c}{\bf\textsc{trf}}&8.1&6.55&10.4&54&75&2.29&3.40&\bf.667&.451&.635&\bf.671&4.7e-4&1.0e-3&\bf.108&.120&.120&.112 \\ \hline
 	
 	\multirow{2}{*}{\bf \rotatebox{90}{ID}} &\textsc{b}&7.4&\bf6.67&\bf10.8&50&69&\bf1.96&\bf2.91&.627&\bf.455&.633&.637&\bf2.1e-3&\bf7.7e-3&.106&.113&.111&.103 \\
 	&\textsc{t}&\bf12.0&6.44&10.4&\bf74&\bf106&2.53&3.79&.646&\bf.456&\bf.637&.651&9.8e-4&3.2e-3&\bf.108&\bf.123&\bf.125&\bf.118\\ \hline
 	
 	\multicolumn{2}{c}{\bf\textsc{RT}}& 13.4&8.26&14.2&113&170&.03&.12&.623&.386&.601&.622&4.6e-2&3.2e-1&.079&.102&.109&.105\\
 	\multicolumn{2}{c}{\bf\textsc{GT}}&13.1&8.18&13.8&110&149&0&0&1&1&1&.655&4.0e-2&3.1e-1&1&1&1&1\\
 	
 \end{tabular}
\end{center}
\caption{\label{table:cornell} Metrics on the unfiltered test set (Cornell) at the validation loss minimum. \textsc{trf} refers to \texttt{transformer}, \textbf{ID} to \textsc{identity}. \textsc{target}-side, and filtering \textsc{both} sides are denoted by initials. Best results are highlighted with bold. \textbf{GT} refers to ground truth responses and \textbf{RT} refers to randomly selected responses from the training set.}
\end{table*}

\begin{table*}[h!]
\begin{center}
 \small
 \renewcommand{\arraystretch}{1.1}
 \begin{tabular}{p{0.1cm}p{0.1cm}p{0.3cm}p{0.4cm}p{0.3cm}p{0.3cm}p{0.3cm}R{0.6cm}R{0.6cm}p{0.3cm}p{0.3cm}p{0.3cm}p{0.4cm}p{0.8cm}p{0.8cm}p{0.5cm}p{0.5cm}p{0.5cm}p{0.5cm}}
 	
 	&& \(|U|\) & \(H_{w}^{u}\) & \(H_{w}^{b}\) & \(H_{u}^{u}\) & \(H_{u}^{b}\) & \(D_{kl}^{u}\) & \(D_{kl}^{b}\) & \textsc{avg} & \textsc{ext} & \textsc{gre} & \textsc{coh} & d1 &d2&b1&b2&b3&b4\\ \hline
 	
 	\multicolumn{2}{c}{\bf\textsc{trf}}&20.6&6.89&\bf11.4&121&177&\bf2.28&\bf3.40&.643&.395&.591&.659&\bf2.1e-3&\bf6.2e-3&.0519&.0666&.0715&.0693 \\ \hline
 	
 	\multirow{2}{*}{\bf \rotatebox{90}{ID}} &\textsc{b}&20.3&\bf6.95&\bf11.4&119&171&2.36&\bf3.41&\bf.657&.394&.595&\bf.673&1.2e-3&3.4e-3&\bf.0563&\bf.0736&.0795&.0774 \\
 	&\textsc{t}&\bf29.0&6.48&10.7&\bf157&\bf226&2.68&3.69&.644&\bf.403&\bf.602&.660&1.4e-3&4.6e-3&\bf.0550&\bf.0740&\bf.0819&\bf.0810\\ \hline
 	
 	\multicolumn{2}{c}{\bf\textsc{RT}}& 14.0&9.81&15.9&136&171&.05&.19&.681&.334&.543&.695&8.5e-2&5.4e-1&.0444&.0751&.0852&.0840\\
 	\multicolumn{2}{c}{\bf\textsc{GT}}&14.0&9.78&15.8&135&167&0&0&1&1&1&.734&8.1e-2&5.3e-1&1&1&1&1\\
 	
 \end{tabular}
\end{center}
\caption{\label{table:twitter} Metrics on the unfiltered test set (Twitter) at the validation loss minimum. \textsc{trf} refers to \texttt{transformer}, \textbf{ID} to \textsc{identity}. \textsc{target}-side, and filtering \textsc{both} sides are denoted by initials. Best results are highlighted with bold. \textbf{GT} refers to ground truth responses and \textbf{RT} refers to randomly selected responses from the training set.}
\end{table*}

Computing the unigram and bigram KL divergence with a uniform distribution instead of the model yields a value of 4.35 and 1.87, respectively. Thus, all models learned a much better distribution, suggesting that this is indeed a useful metric. We believe the main reason that clustering methods perform worse than
\textsc{identity} is that clustering adds some noise to the filtering process. Conducting a good clustering of sentence vectors is a hard task.
This could be remedied by filtering only utterances instead of whole clusters, thus combining \textsc{identity} and the clustering methods. In this scenario, the entropy of individual utterances is computed based on the clustered data. The intuition behind this approach would be that
the noise in the clusters based on which we compute entropy is less harmful than the noise in clusters which we consider for filtering. Finally, Table~\ref{table:examples} shows responses from the baseline and the best performing model to 3 randomly selected inputs from the test set (which we made sure are not present in the training set) to show that training on filtered data does not degrade response quality. We show more 
example responses in Appendix~\ref{ssec:eval_appendix}.

\subsection{Cornell and Twitter Results}
\label{ssec:other_datasets}
To further solidify our claims we tested the two best performing variants of \textsc{identity} (\textsc{both} and \textsc{target}) on the Cornell Movie-Dialogs Corpus and on a subset of 220k examples from the Twitter corpus\footnote{\url{https://github.com/Marsan-Ma/chat_corpus/}}. Entropy thresholds were selected to be similar to the DailyDialog experiments (Table~\ref{table:thresholds}). Evaluation results at the validation loss minimum on the Cornell corpus and the Twitter dataset are presented in Table~\ref{table:cornell} and Table~\ref{table:twitter}, respectively. On these noisier datasets our simple \textsc{identity} method still managed to improve over the baseline, but the impact is not as pronounced and in contrast to DailyDialog, \textsc{both} and \textsc{target} perform best on nearly the same number of metrics. On these noisier datasets the clustering methods might work better, this is left for future work. Compared to DailyDialog there are some important distinctions that also underline that these datasets are of lesser quality. The \textsc{coherence} metric is worse on the ground truth responses than on model responses (Table~\ref{table:cornell}, and some embedding metrics and BLEU scores are better on randomly selected responses than on model responses (Table~\ref{table:twitter}).

\section{Conclusion}
\label{sec:conclusion}
We proposed a simple unsupervised entropy-based approach that can be applied to 
any conversational dataset for filtering generic
sources/targets that cause ``confusion'' during the training of 
open-domain dialog models. We compared various setups in an extensive 
quantitative evaluation, and showed that the best approach is measuring the entropy of individual utterances and filtering pairs based on the entropy of target, but not source utterances. Some limitations of current automatic 
metrics and the loss function have also been shown, by examining their behavior 
on random data and with overfitting.

In the future, we plan to explore several additional ideas. As mentioned in 
Section~\ref{ssec:results}, we want to extend our clustering experiments combining the ideas behind \textsc{identity} and the clustering methods to make them more robust to noise. We wish to conduct clustering experiments on noisier datasets and try other sentence representations \cite{Devlin:2018}. We also plan to combine our method with coherence-based filtering \cite{Xu:2018}. 
Furthermore, we intend to perform a direct quantitative evaluation of our method based on human evaluation. Finally, we believe our method is general enough 
that it could also be applied to datasets in other similar NLP tasks, such as 
machine translation, which could open another interesting line of future research.

\section*{Acknowledgments}
We wish to thank Evelin \'{A}cs, P\'{e}ter Ih\'{a}sz, M\'{a}rton Makrai, Luca Szegletes, and all anonymous reviewers for their thoughtful feedback.
Work partially supported by Project FIEK 16-1-2016-0007, financed by the FIEK\_16 funding scheme of the Hungarian National Research, Development and Innovation Office (NKFIH). 

\bibliography{../../Common/Bib/ml}
\bibliographystyle{../../Common/ACL/ACL19/acl_natbib}

\clearpage
\appendix
\section{Appendix}
\subsection{High Entropy Utterances}
\label{ssec:high_entropy}
\subsubsection{Top 20 high entropy utterances}
\begin{table}[htb!]
\begin{center}
 \begin{tabular}{lll}
 	Utterance & Frequency & Entropy \\ \hline
 	yes  .  & 173 & 7.06 \\
 	thank you  .  & 141 & 6.57 \\
 	why  ?  & 104 & 6.33 \\
 	here you are  .  & 99 & 6.10 \\
 	ok  .  & 75 & 6.00 \\
 	what do you mean  ?  & 77 & 5.97 \\
 	may i help you  ?  & 72 & 5.96 \\
 	can i help you  ?  & 80 & 5.93 \\
 	really  ?  & 74 & 5.91 \\
 	sure  .  & 66 & 5.66 \\
 	what can i do for you  ?  & 51 & 5.63 \\
 	why not  ?  & 61 & 5.42 \\
 	what  ?  & 48 & 5.27 \\
 	what happened  ?  & 44 & 5.18 \\
 	anything else  ?  & 43 & 5.17 \\
 	thank you very much  .  & 72 & 5.14 \\
 	what is it  ?  & 41 & 5.06 \\
 	i see  .  & 42 & 5.05 \\
 	no  .  & 42 & 5.04 \\
 	thanks  .  & 50 & 5.03 \\
 	
 \end{tabular}
\end{center}
\caption{\label{table:high_entropy_utts} Top 20 source utterances (from DailyDialog) sorted by entropy. The entropy was calculated with \textsc{identity}.}
\end{table}

\subsubsection{High Entropy Clusters}
\begin{figure}[h!]
\centering
\includegraphics[width=0.47\textwidth]{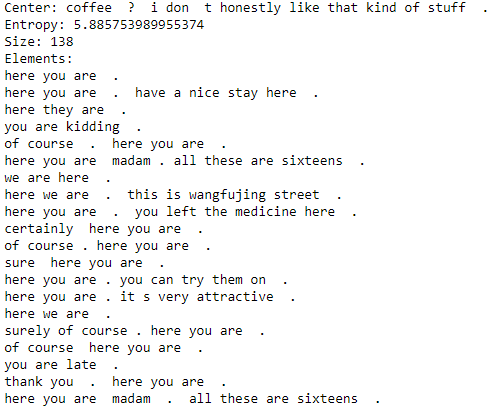}
\caption{A high entropy cluster from DailyDialog.}
\label{fig:cluster1}
\end{figure}

\begin{figure}[h!]
\centering
\includegraphics[width=0.47\textwidth]{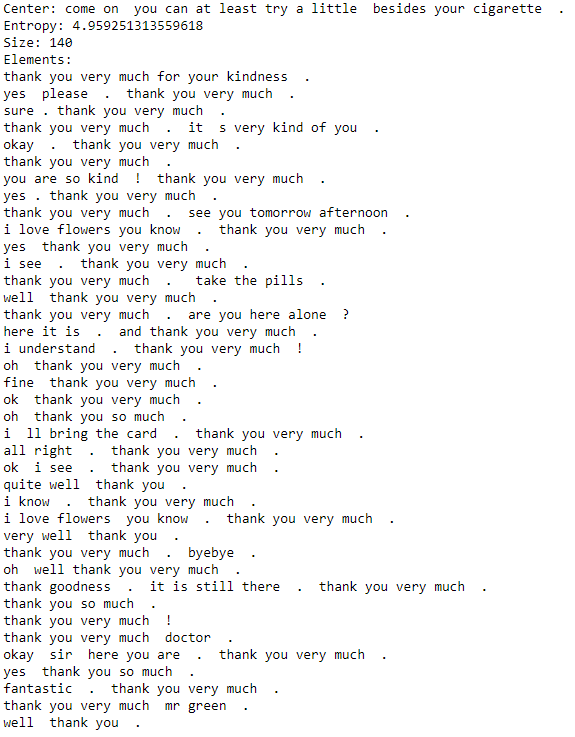}
\caption{A high entropy cluster from DailyDialog.}
\label{fig:cluster2}
\end{figure}

\begin{figure}[h!]
\centering
\includegraphics[width=0.47\textwidth]{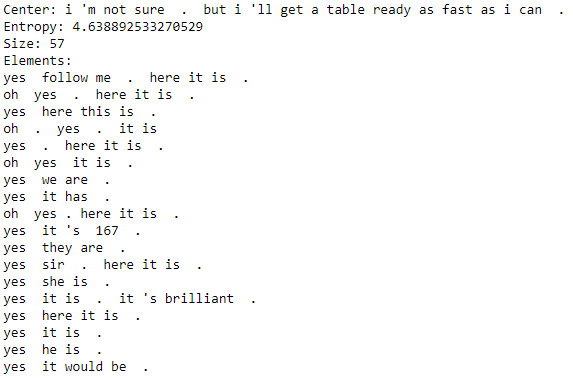}
\caption{A high entropy cluster from DailyDialog.}
\label{fig:cluster3}
\end{figure}

\clearpage
\subsection{Model Parameters}
\label{ssec:parameters_appendix}

\begin{table}[h!]
\begin{center}
 \begin{tabular}{ll}
 	Name & Value \\ \hline
 	
 	Hidden size & 512 \\
 	Number of hidden layers & 6 \\
 	Label smoothing & 0.1 \\
 	Filter size & 2048 \\
 	Number of attention heads & 8 \\
 	Layer dropout & 0.2 \\
 	Relu dropout & 0.1 \\
 	Attention dropout & 0.1 \\
 	Learning rate & 0.2 \\
 	Learning rate warmup steps & 8000 \\
 	
 \end{tabular}
\end{center}
\caption{\label{table:transformer params} Transformer hyperparameters.}
\end{table}

\subsection{Evaluation Metrics and Examples}
\label{ssec:eval_appendix}

\begin{figure}[h!]
\centering
\includegraphics[width=0.47\textwidth]{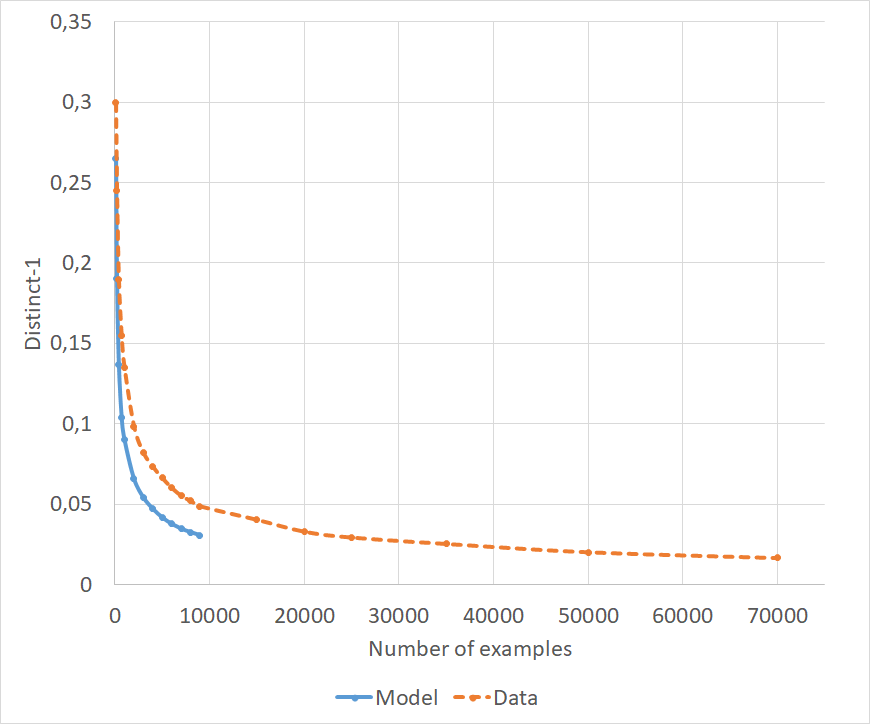}
\caption{Distinct-1 metric with respect to number of test examples (on DailyDialog). Model responses were evaluated on 9000 examples only, since the rest were training examples.}
\label{fig:distinct1}
\end{figure}
\begin{figure}[h!]
\centering
\includegraphics[width=0.47\textwidth]{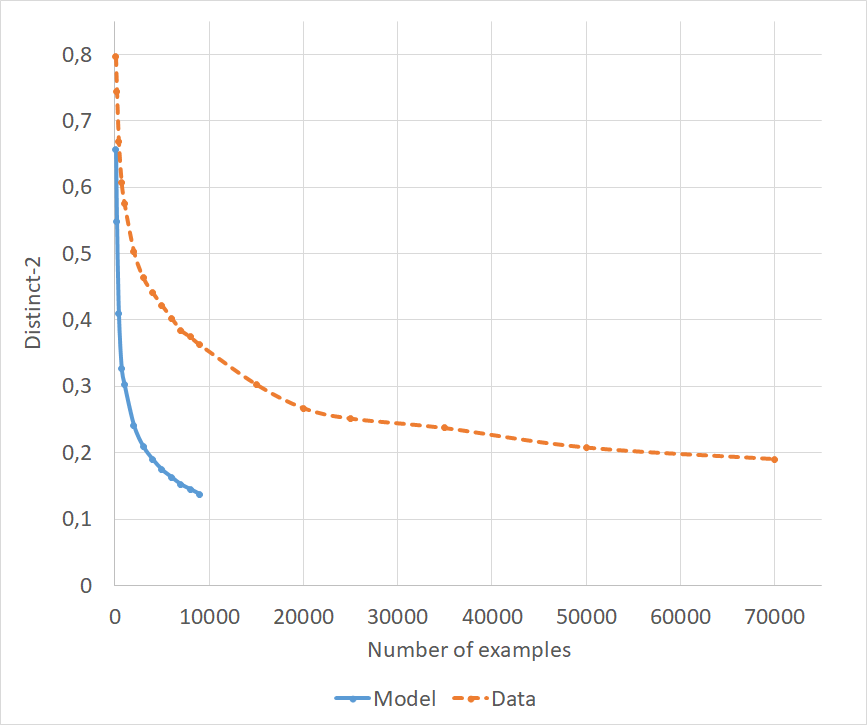}
\caption{Distinct-2 metric with respect to number of test examples (on DailyDialog). Model responses were evaluated on 9000 examples only, since the rest were training examples.}
\label{fig:distinct2}
\end{figure}

\begin{figure*}[h!]
\centering
\includegraphics[width=0.95\textwidth]{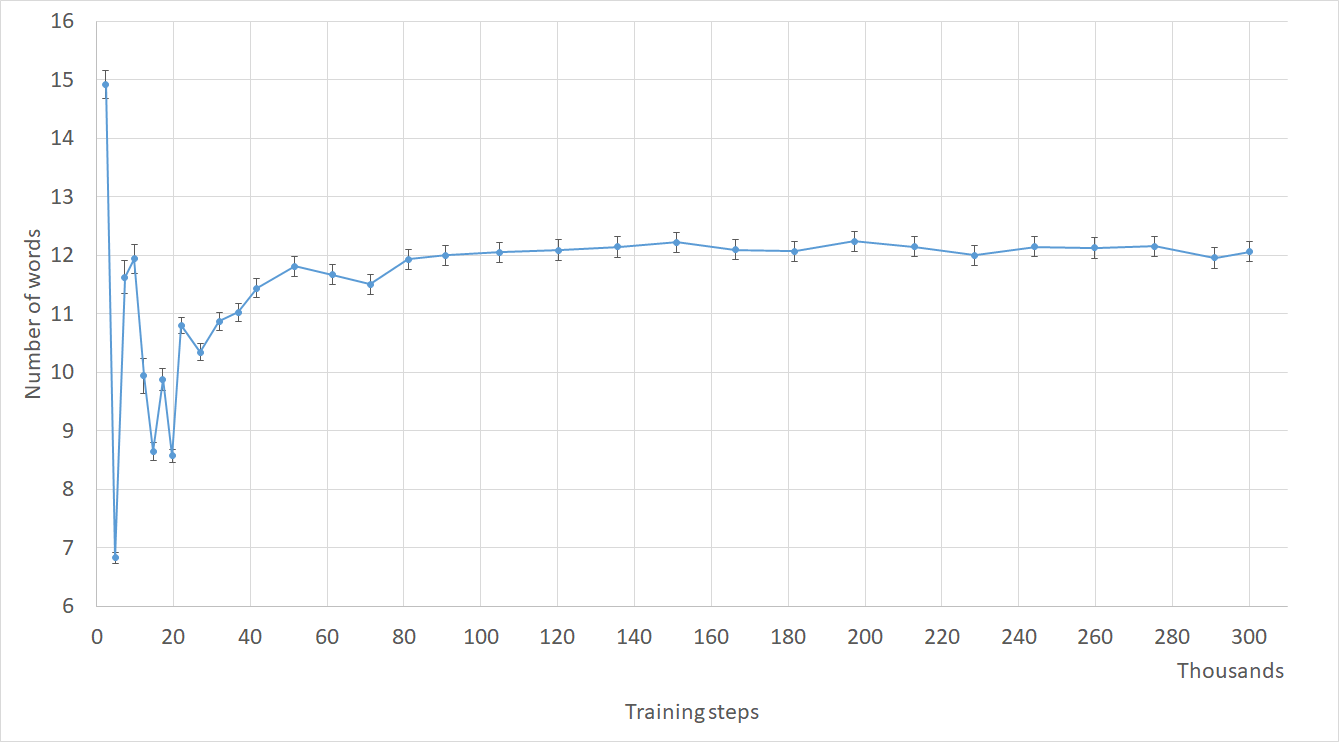}
\caption{Average length of responses (computed on the validation set) with respect to the number of training steps of the \texttt{transformer} trained on unfiltered data (DailyDialog).}
\label{fig:length_base}
\end{figure*}
\begin{figure*}[h!]
\centering
\includegraphics[width=0.95\textwidth]{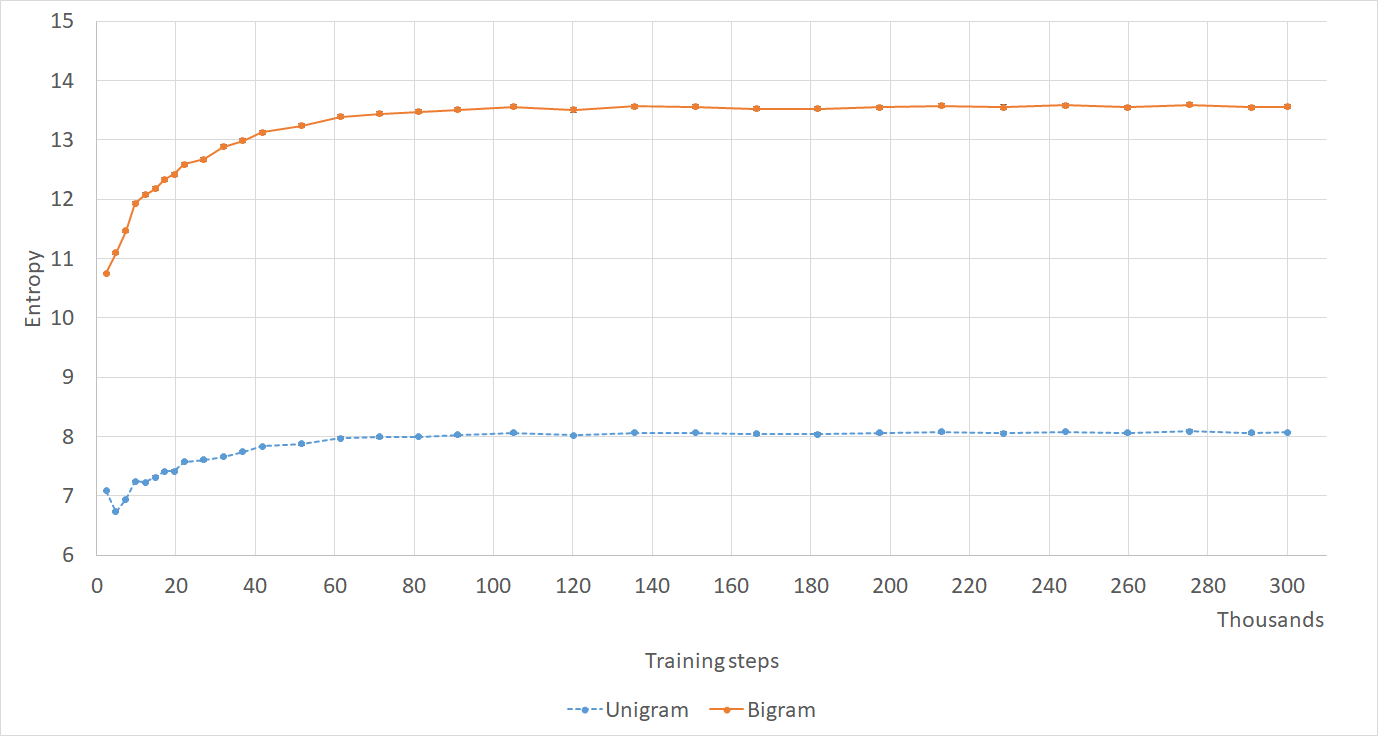}
\caption{Word entropy of responses (computed on the validation set) with respect to the number of training steps of the \texttt{transformer} trained on unfiltered data (DailyDialog).}
\label{fig:entropy_base}
\end{figure*}
\begin{figure*}[h!]
\centering
\includegraphics[width=0.95\textwidth]{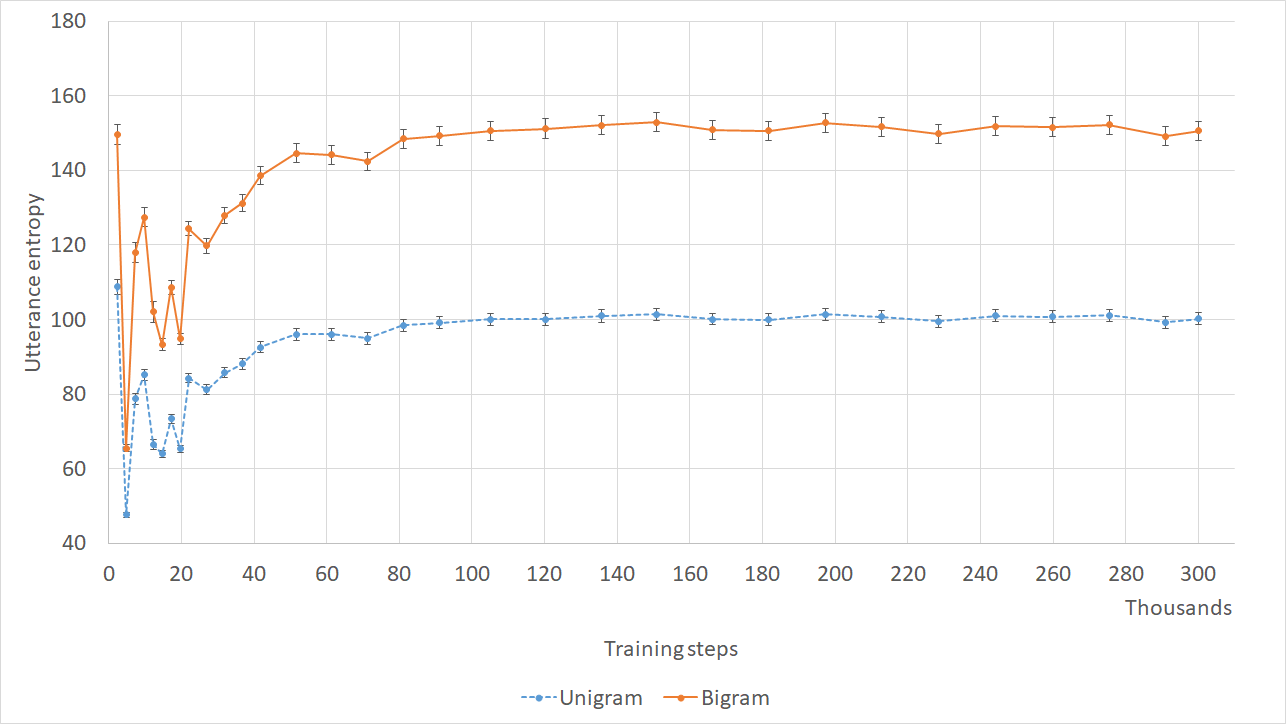}
\caption{Utterance entropy of responses (computed on the validation set) with respect to the number of training steps of the \texttt{transformer} trained on unfiltered data (DailyDialog).}
\label{fig:utt_entropy_base}
\end{figure*}

\begin{figure*}[h!]
\centering
\includegraphics[width=0.95\textwidth]{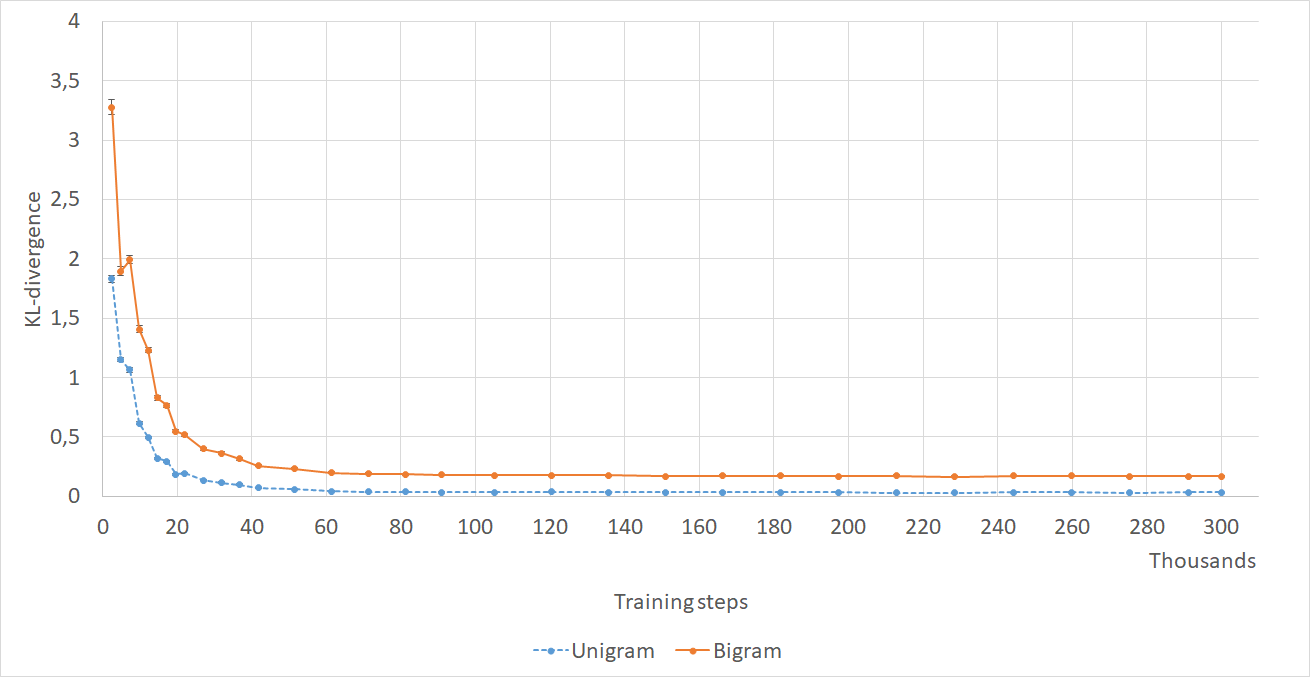}
\caption{KL divergence of responses (computed on the validation set) with respect to the number of training steps of the \texttt{transformer} trained on unfiltered data (DailyDialog).}
\label{fig:divergence_base}
\end{figure*}
\begin{figure*}[h!]
\centering
\includegraphics[width=0.95\textwidth]{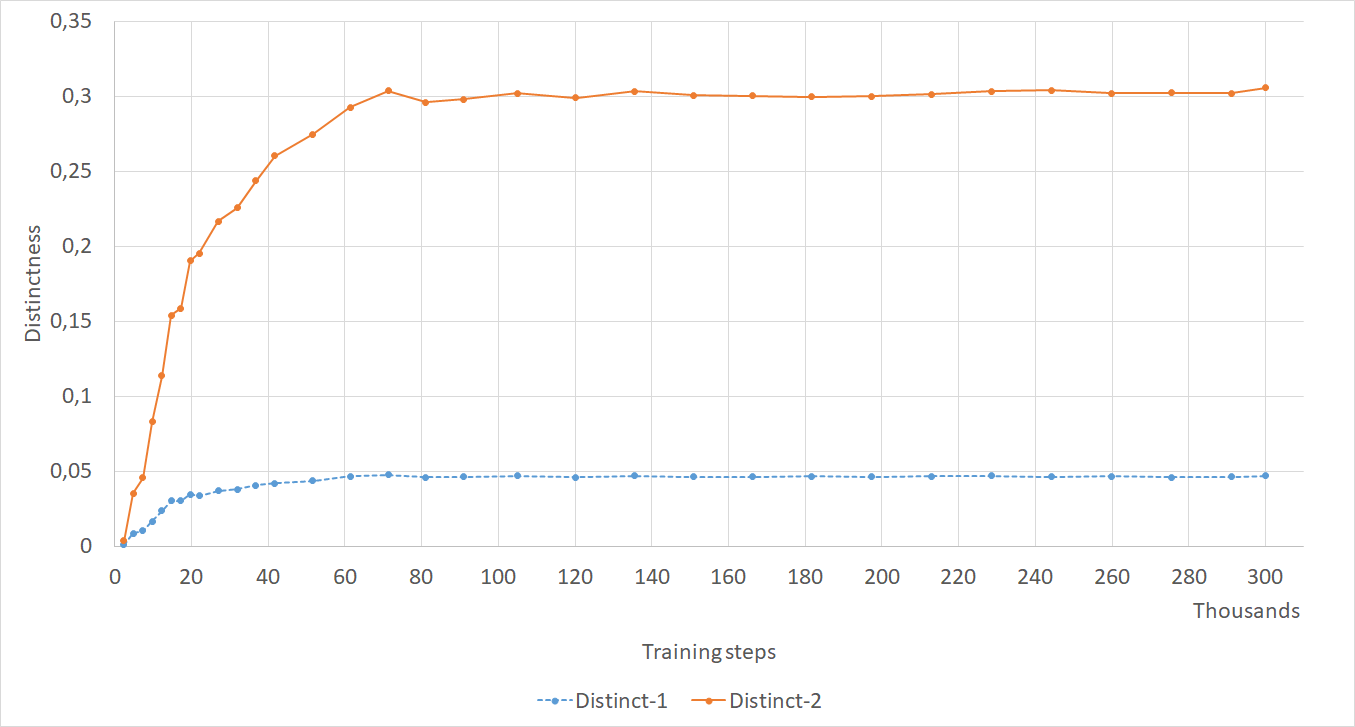}
\caption{Distinct-1 and distinct-2 metrics (computed on the validation set) with respect to the number of training steps of the \texttt{transformer} trained on unfiltered data (DailyDialog).}
\label{fig:distinct_base}
\end{figure*}

\begin{table*}[h!]
\begin{center}
 \fontsize{8}{12}\selectfont
 \renewcommand{\arraystretch}{0.8}
 \begin{tabular}{p{2.7cm}p{2.7cm}p{2.7cm}p{2.7cm}p{2.7cm}}
 	
 	\bf Input & \bf Unfiltered &\bf Unfiltered-Overfitted & \bf \textsc{identity} \textsc{target} filtered & \bf \textsc{identity} \textsc{target} filtered-overfitted \\ \hline
 	
 	some high speed chase scene i think .&i 'm not sure .&but many things are went wrong !&i 'm so sorry .&enough !\\ \hline
 	how long is this bus ride ?&about twelve hours .&it 's about fortyfive minutes to an hour .&it 's about 20 minutes .&it 's only about fortyfive minutes to an hour .\\ \hline
 	is that yellow bus it ?&no it 's not far . it 's a bit cold .&yeah . there s a new one in america .&no it 's a yellow one .&no it 's not very expensive these days .\\ \hline
 	hi mary ! what do you like to do in your spare time ?&i like playing basketball .&well i like to play the piano .&i like playing basketball .&i like playing chess .\\ \hline
 	let me see . we can visit the country . we have n't done that for such a long time .&i hope so .&yes there are many famous paintings on show . i hope it 's the nicest city in the united states .&i 'm glad to hear that .&where are your parents ?\\ \hline
 	thirty u . s . dollars . . . and it 's only been used for six months .&that 's fine .&i see . is it a long time to come to the station ?&how many months did you get there ?&what 're two weeks ?\\ \hline
 	five hundred dollars altogether .&here you are .&here you are .&here is your bill .&here you are . thank you byebye .\\ \hline
 	look at this bridge . we will soon be crossing the thames .&yes i agree .&oh yeah .&how far is it ?&well there 's a spanish teacher . so i 'll often go back to the spanish museum .\\ \hline
 	is the furniture crafted well ?&yes it is .&yes it 's fully furnished .&yes the furniture is right over there .&yes it 's a furnished eatin kitchen . the furniture and the icebox are all pretty outdated .\\ \hline
 	where do they go ?&they 're on the bottom floor .&i have no idea .&they re in the cabinet next to the library .&go to them department please .\\ \hline
 	let me pay for it .&ok . here you are .&cash credit card or debit card ?&ok . here is the money .&cash credit card or debit card ?\\ \hline
 	he never does and in fact he seldom communicates with him .&he is a hard job . he can also learn a lot from the government .&that 's right . he has no children and his wife experience too .&he is a very kind of man .&yeah . he 's a very easy teacher .\\ \hline
 	so you must have had a busy and enjoyable day yesterday .&yes i am . i really enjoyed it .&yes . i am very sorry i can t make it .&i will . thanks for coming .&yes . let 's go .\\ \hline
 	yes . the reservation under mr . watson .&ok .&could you spell that please ?&i 'm afraid there 's no seat in the middle of the table .&ok . would you please fill out the table ?\\ \hline
 	there are so many to choose from . i did n't know you could do all of this using a card .&i see .&what 's the best part ?&i 'm not sure . i 'll go to the doctor and get my card right away .&you mean you do n't have to tell me the truth ?\\ \hline
 	it was great ! how was your year ?&it was great . i had a great time .&my mother had a birthday party for me .&it was great . i 'm glad i did n't like it .&it was ok . there was a lot of turbulence though . the worst thing was n't too good !\\ \hline
 	tomorrow ? no i am afraid i won t be able to . i m going on business to birmingham .&i think you are looking for a .\textless unk\textgreater \textless unk\textgreater in the future . the best reason why are you want to see the results .&well there is a problem with the exchange rate . it s the same size that you have in the u . s .&you should go to the dentist .&you might want to change your plans . with the cold and wet weather that we re supposed to have tomorrow the roads are going to be rather slippery .\\

 \end{tabular}
\end{center}
\caption{\label{table:responses} Responses to randomly selected test inputs which we made sure were not in the training data (DailyDialog). \textit{Unfiltered} is the model trained on unfiltered data, and \textsc{identity} \textsc{target} is the model trained on \textsc{identity}, \textsc{target} filtered data. \textit{Overfitted} means that the respective model is evaluated at an overfitted point.}
\end{table*}

\begin{table*}[h!]
\begin{center}
 \fontsize{8}{12}\selectfont
 \renewcommand{\arraystretch}{0.8}
 \begin{tabular}{p{2.7cm}p{2.7cm}p{2.7cm}p{2.7cm}p{2.7cm}}
 	
 	\bf Input & \bf Unfiltered &\bf Unfiltered-Overfitted & \bf \textsc{identity} \textsc{target} filtered & \bf \textsc{identity} \textsc{target} filtered-overfitted \\ \hline
 	
 	you 're flattering me .&no .&no really .&no but you can t .&no really . i can only hear a little accent .\\ \hline
 	three bags and a suitcase . this is my luggage to check .&ok . i will take one .&what 's this ? essential oil ? wow ! this is a small bag .&i 'll put you in the bag .&ok . is this your luggage properly labelled sir ?\\ \hline
 	i like orange better .&what 's your favorite flavor ?&how about drinks ?&i 'll tell you what you 're doing .&could i have some water ?\\ \hline
 	i like to watch barbie sticking her tongue out .&what 's her figure ?&oh she 's a pretty woman .&you are not supposed to be serious .&oh shut up ! she 's not what you are thinking about .\\ \hline
 	if you give us your inquiry i shall go very carefully into the price and try my best to put you on the best of the terms .&ok . i will take it .&thank you very much . we will try our best to do our best to help you .&i understand . i understand that you are right .&ok i will . i look forward to it .\\ \hline
 	ok . any time on friday will be ok with me .&ok . see you then .&i think so .&i 'll be expecting you then .&great !\\ \hline
 	no wonder you can control your voice so well . you are a professional singer .&i am not a singer .&i would n't mind a position that would be great .&i am a singer . i think i can be a singer in the lead .&i would n't like to be a singer but i love to sing that dream .\\ \hline
 	when can i get high speed internet installed ?&you can make a list of the company and make sure that you do n't have to use the machine .&you 'll need to make an appointment before the company starts .&you can get up at the airport and you 'll have to make an appointment .&you 'll have to make an appointment before we can come in and install it .\\ \hline
 	i like those kinds of programmes too . they re very informative . i think that many people underrate the education value of tv .&i agree . people often criticize tv for showing too much sex and violence .&i agree . people often criticize tv for showing too much sex and violence .&i think that teenagers should be often become addicted and violence .&i agree . people often criticize tv for showing too much sex and violence .\\ \hline
 	can you tell that i m excited ?&i m afraid i m not a child .&i m a fashion designer . next month i have a fashion show .&i m not sure . what do you want to know ?&i m sure that i ll cook the same for you .\\ \hline
 	would you like to have a shampoo sir ?&yes i 'd like a glass of iced tea .&no thank you .&no i 'd rather have it with me .&yes with the lux dandruff shampoo . it helps a lot to get rid of my dandruff .\\ \hline
 	what else would i prepare sir ?&you can bring your beef with your chicken and a bottle of wine .&oh good ! can you make some recommendation for me ?&well the interviewers are not interested in the position .&let me see . everything looks fine .\\ \hline
 	he always says i am a hard worker with consciousness of responsibility sufficient education and enough experience .&i think that maybe you are right .&what s the matter with you ?&how many years of experience will you earn ?&would you like to work in the finance department ?\\ \hline
 	what made you think that ?&i like it a lot .&i love that shirt on you .&i think it 's great . i think i ve learned a lot from different countries .&i will care for it .\\ \hline
 	i can tell you what bus to catch but you have to walk a little bit .&i do n't know .&tell me how to get a ticket and a student bus station .&i 'm sorry but i do n't have to wait .&you 're going to have a car .\\

 \end{tabular}
\end{center}
\caption{\label{table:responses2} Responses to randomly selected test inputs which we made sure were not in the training data (DailyDialog). \textit{Unfiltered} is the model trained on unfiltered data, and \textsc{identity} \textsc{target} is the model trained on \textsc{identity}, \textsc{target} filtered data. \textit{Overfitted} means that the respective model is evaluated at an overfitted point.}
\end{table*}

\end{document}